\documentclass[11pt,a4paper]{article}
\usepackage[hyperref]{acl2020}
\usepackage{times}
\usepackage{latexsym}

\usepackage{microtype}

\usepackage{amsmath,amsfonts,amssymb}
\usepackage{graphicx,multirow}
\usepackage{enumitem}
\usepackage{latexsym, url}
\usepackage{algorithm}
\usepackage{algpseudocode}
\usepackage{colortbl} 
\usepackage{arydshln} 

\DeclareMathOperator*{\argmax}{argmax}

\makeatletter
\newcommand{\thickhline}{%
	\noalign {\ifnum 0=`}\fi \hrule height 1pt
	\futurelet \reserved@a \@xhline
}
\newcolumntype{"}{@{\hskip\tabcolsep\vrule width 1pt\hskip\tabcolsep}}
\makeatother

\aclfinalcopy 

\title{Learning Goal-oriented Dialogue Policy with Opposite Agent Awareness}

\author{
	Zheng Zhang$^1$, Lizi Liao$^2$, Xiaoyan Zhu$^1$, Tat-Seng Chua$^2$,  Zitao Liu$^3$, Yan Huang$^3$, Minlie Huang$^1$\footnotemark[1]\\
	$^1$ DCST, Tsinghua University, Beijing, China \\
	$^2$ School of Computing, National University of Singapore, Singpore \quad
	$^3$ TAL AI Lab \\
	$^1${\small \tt zhangz.goal@gmail.com \quad \{zxy-dcs,aihuang\}@tsinghua.edu.cn} \\
	$^2${\small \tt liaolizi.llz@gmail.com \quad chuats@comp.nus.edu.sg} \\
	$^3${\small \tt zitao.jerry.liu@gmail.com \quad galehuang@100tal.com}
}

\date{}

\begin{document}
	\maketitle
	\renewcommand{\thefootnote}{\fnsymbol{footnote}}
	\footnotetext[1]{Corresponding author.} 
	
	\begin{abstract}
		Most existing approaches for goal-oriented dialogue policy learning used reinforcement learning, which focuses on the target agent policy and simply treat the opposite agent policy as part of the environment. While in real-world scenarios, the behavior of an opposite agent often exhibits certain patterns or underlies hidden policies, which can be inferred and utilized by the target agent to facilitate its own decision making. This strategy is common in human mental simulation by first imaging a specific action and the probable results before really acting it. We therefore propose an opposite behavior aware framework for policy learning in goal-oriented dialogues. We estimate the opposite agent's policy from its behavior and use this estimation to improve the target agent by regarding it as part of the target policy. We evaluate our model on both cooperative and competitive dialogue tasks, showing superior performance over state-of-the-art baselines.
	\end{abstract}

	\section{Introduction}

In goal-oriented dialogue systems, dialogue policy plays a crucial role by deciding the next action to take conditioned on the dialogue state.
This problem is often formulated using reinforcement learning (RL) in which the user serves as the environment \cite{levin1997learning, young2013pomdp, fatemi2016policy, zhao2016towards, dhingra2016towards, su2016continuously, li2017end, williams2017hybrid, liu2017iterative, lipton2018bbq, liu2018dialogue, gao2019neural, takanobu2019guided}.
However, different from symbolic-based and simulation-based RL tasks, such as chess \cite{silver2016mastering} and video games \cite{mnih2015human}, which can get vast amounts of training interactions in low cost, dialogue system requires to learn directly from real users, which is too expensive.

Therefore, there are some efforts using simulation methods to provide an affordable training environment.
One principle direction for mitigating this problem is to leverage human conversation data to build a user simulator, and then to learn the dialogue policy by making simulated interactions with the simulator \cite{schatzmann2006survey, li2016user, gur2018user}.
However, there always exist discrepancies between simulated users and real users due to the inductive biases of the simulation model, which can lead to a sub-optimal dialogue policy \cite{dhingra2016towards}.
Another direction is to learn the dynamics of dialogue environment during interacting with real user, and concurrently use the learned dynamics for RL planning\cite{peng2018deep, su2018discriminative, wu2018switch, zhang2019budgeted}. Most of these works are based on Deep Dyna-Q (DDQ) framework \cite{sutton1990integrated}, where a world model is introduced to learn the dynamics (which is much like a simulated user) from real experiences. The target agent's policy is trained using both real experiences via direct RL and simulated experiences via a world-model.

\begin{figure*}
	\centering
	\includegraphics[width=1.0\linewidth]{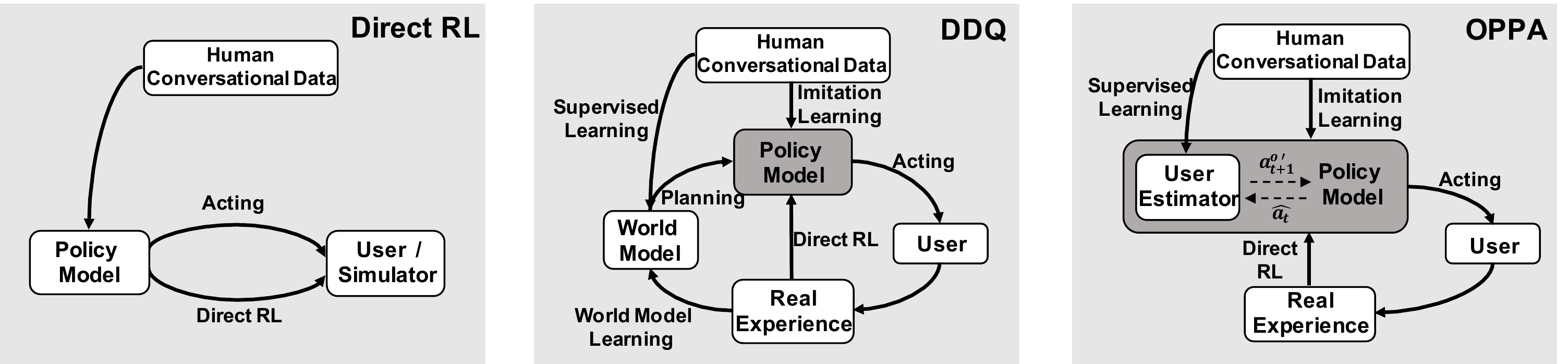}
	\caption{A comparison of dialogue policy learning a) with real/simulated user, b) with real user via DDQ and c) with real user guided by active user estimation.}
	\label{fig:framework}
\end{figure*}

In the above methods, both the simulated user and world model facilitate target policy learning by providing more simulated experiences and remain a black box for the target agent. That is, the target agent's knowledge about the simulated agents is still passively obtained through interaction and implicitly learned by the policy model updating as in direct try-and-error with real user.
However, we argue that from the angle of a target agent, actively exploring the world with proper estimation would not only make user simulation more efficient but also improve the target agent's performance. In agreement with the findings from cognitive science, humans often maintain models of other people they interact with to capture their goals \cite{C14-1001,premack1978does}. And humans manage to use their mental process to simulate others' behavior \cite{gordon1986folk,gallese1998mirror}. Therefore, to carefully treat and model the behaviors of other agents would be a way full of potential. For example, in competitive tasks such as chess, the player often see a number of moves ahead by considering the possible reaction of the other player. In goal-oriented dialogues for a hotel booking task, the agent can reduce interaction numbers and improve user experience by modeling users as business traveler with strict time limit or backpacker seeking adventure.

In this paper, we propose a new dialogue policy learning method with OPPosite agent Awareness (OPPA), where the agent maintains explicit modeling of the opposite agent or user for facilitating its own policy learning. Different from DDQ, the estimated user model is not utilized as a simulator to produce simulated experiences, but as an auxiliary component of the target agent's policy to guide the next action. Figure \ref{fig:framework}(c) shows the framework of our model. Specifically, whenever the system needs to output an action, it foresees a candidate action $\hat{a_t}$ and consequently estimates the user's response behavior $a_{t+1}^o{'}$. On top of this estimation, as well as the dialogue context, it makes better decisions with a dynamic estimation of the user's strategy. To further regulate the behavior of the system agent, we mitigate the difference between the real system action $a_t$ and the sampled action $\hat{a_t}$ with decay for better robustness and consistency. Without any constraint on the type of agents (either competitive or cooperative), the proposed OPPA method can be applied to both cooperative and non-cooperative goal-oriented dialogues.



To summarize, our contributions are three-fold:
\vspace{-0.3cm}
\begin{itemize}
    \item We propose a new dialogue policy learning setting where the agent shifts from passively learning to actively estimating the opposite agent or user for more efficient simulation, thereby obtaining better performance. \vspace{-0.2cm}
	\item We mitigate the difference between real system agent action and the sampled action with decay to further enhance estimated system agent behavior. \vspace{-0.2cm}
	\item Extensive experiments on both cooperative and competitive goal-oriented dialogues indicate that the proposed model can achieve better dialogue policies than baselines.
\end{itemize}

	\section{Related Work}

\subsection{RL-based Dialogue Policy Learning}
Policy learning plays a central role in building a goal-oriented dialogue system by deciding the next action, which is often formulated using the RL framework.
Early methods used probabilistic graph model, such as partially observable Markov decision process (POMDP), to learn dialogue policy by modeling the conditional dependences between observation, belief states and actions \cite{williams2007partially}. However, these methods require manual work to define features and state representation, which leads to poor domain adaptation.
More recently, deep learning methods are applied in dialogue policy learning, including DQN \cite{mnih2015human} and Policy Gradient \cite{sutton2000policy} methods, which mitigate the problem of domain adaptation through function approximation and representation learning \cite{zhao2016towards}.

Recently, there are some efforts focused on multi-domain dialogue policy.
An intuitive way is to learn independent policies for each specific domain and aggregate them \cite{wang2014policy,gavsic2015policy,cuayahuitl2016deep}. 
There are also some works using hierarchical RL,  which decomposes the complex task into several sub-tasks \cite{peng2017composite,casanueva2018feudal} according to pre-defined domain structure and cross-domain constraints.
Nevertheless, most of the above works regard the opposite agent as part of the environment without explicitly modeling its behavior.

Planning based RL methods are also introduced to make a trade-off between reducing human interaction cost and learning a more realistic simulator. \cite{peng2018deep} proposed to use Deep Dynamic Q-network, in which a world model is co-trained with the target policy model.
By training the world model with the real system-human interaction data, it consistently approaches the performance of real users, which provides better simulated experience for planning. Adversarial methods are applied to dynamically control the proportion of simulated and real experience during different stages of training \cite{su2018discriminative,wu2018switch}. Still, these methods work from the opposite agents' angle. 

\vspace{-2mm}
\subsection{Dialogue User Simulation}
In RL-based dialogue policy learning methods, a user simulator is often required to provide affordable training environments due to the high cost of collecting real human corpus.
Agenda-based simulation \cite{schatzmann2007agenda,li2016user} is a widely applied rule-based method, which starts with a randomly generated user goal that is unknown to the system. During a dialogue session, it remains a stack data structure known as \textit{user agenda}, which holds some pending user intentions to achieve. In the stack update process, machine learning or expert-defined methods can be applied.
There are also some model-based methods that learn a simulator from real conversation data. Recently, seq2seq framework is introduced by encoding dialogue history and generates the next response or dialogue action \cite{asri2016sequence,kreyssig2018neural}. By incorporating a variational step to the seq2seq network, it can introduce meaningful diversity into the simulator \cite{gur2018user}. Our work tackles the problem from a different point of view. We let the target agent approximate an opposite agent model to save user simulation efforts.

	\section{Model}
\label{sec:model}
In this section, we introduce our proposed OPPA model. There are two agents in our framework, one is the system agent we want to optimize, and the other is the user agent. We refer to these two agents as \textit{target} and \textit{opposite} agents in the following sections. Note that the proposed model works at dialog act level, and it can also work at natural language level when equipped with natural language understanding (NLU) and natural language generation (NLG) modules.


\subsection{Overview}
\begin{figure*}
	\centering
	\includegraphics[width=0.88\linewidth]{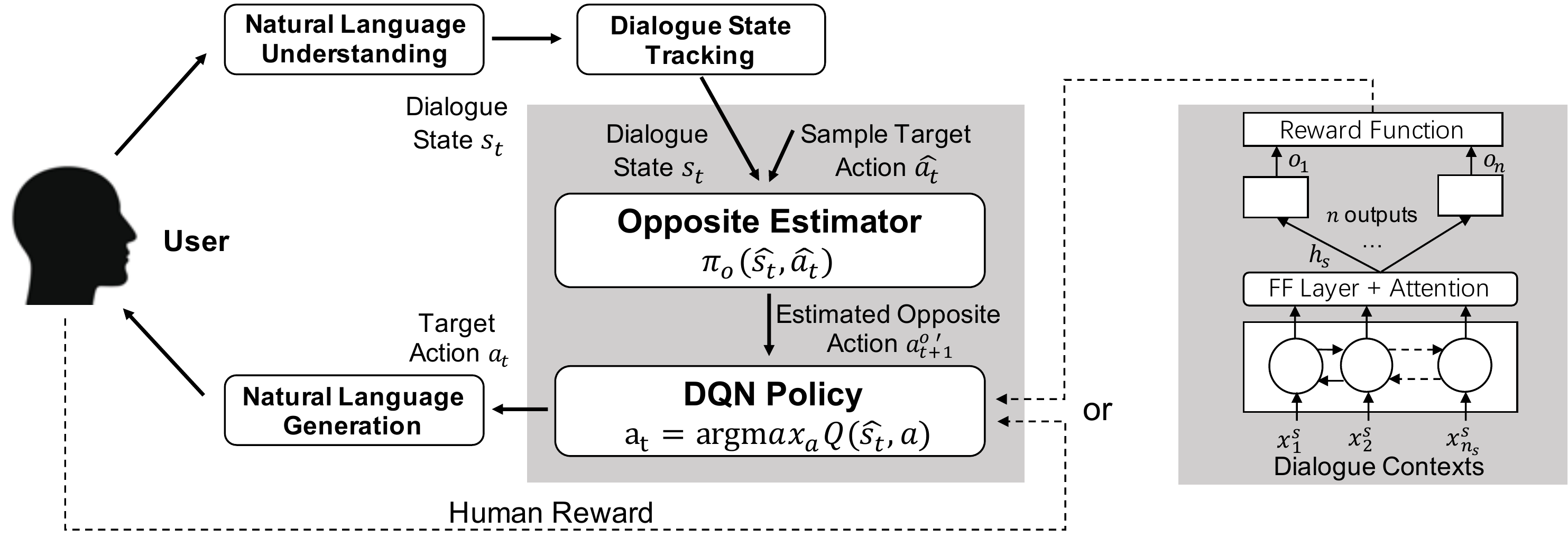}
	\caption{The proposed OPPA model and the reward function. Note that the reward for policy model can be either from real user or the reward function depending on whether real reward is available.}
	\label{fig:model}
\end{figure*}

As shown in Figure \ref{fig:model}, the proposed model consists of two key components: a target agent Q-function $Q(s, a)$ and an opposite agent policy estimator $\pi^o(s, a)$. 
Specifically, each time before the target agent needs to take an action, the model samples a candidate action $\hat{a_t}$. Then the opposite estimator $\pi_o$ estimates the opposite agent's response behavior $a_{t+1}^o{'}$, which is then aggregated with the original dialog state $s_t$ to generate a new state $\hat{s_t}$. On top of this new state, the target policy $Q(s, a)$ gets the next target action. In more detail, a brief script of our proposed OPPA model is shown in Algorithm \ref{algo}.

\begin{algorithm}[ht]
	\normalsize
	\caption{OPPA for Dialogue Policy Learning}\label{algo}
	\begin{algorithmic}[1]
		\Require $\epsilon$, $C$
		\State initialize $\pi^o(s,a;\theta_\pi)$ and $Q(s, a;\theta_Q)$ by supervised and imitation learning
		\State initialize $Q'(s,a,\theta_{Q'})$ with $\theta_{Q'}=\theta_Q$
		\State initialize replay buffer $D$
		\For{\textbf{each} \textit{iteration}}
		\State $user$ acts $a^u$
		\State initialize state $s$
		\While{not $done$}
		\State $e = $ random$(0, 1)$
		\If{$e < \epsilon$}
		\State select a random action $a$
		\Else
		\State sample $\hat{a_t}$
		\State est. user action ${a_{t+1}^o}'=\pi^o(s, \hat{a_t})$
		\State $\hat{s} = [s, E^o{a_{t+1}^o}']$
		\State $a = $ argmax$_{a'}Q(\hat{s}, a';\theta_Q)$
		\EndIf
		\State execute $a$
		\State get user response $a^o$ and reward $r$
		\State state updated to $s'$
		\State store $(s, a, r, s')$ to $D$
		\EndWhile
		\State sample minibatches of $(s, a, r, s')$ from $D$
		\State update $\theta_Q$ according to Equation \ref{eq:dqn}
		\State each every $C$ iterations set $\theta_{Q'}=\theta_Q$
		\EndFor
	\end{algorithmic}
\end{algorithm}

\subsection{Opposite Action Estimation}
One essential target of the opposite estimator is to measure how the opposite agent reacts given its preceding target agent action and state. In OPPA, we implement the opposite estimation model using a two-layer feed-forward neural network followed by a softmax layer. It takes as input the current state $s_t$, a sampled target action $\hat{a_t}$, and predicts an opposite action $a_{t+1}^o{'}$ as below:
\begin{equation}
a_{t+1}^o{'}=\pi^o(s_t,\hat{a_t}).
\end{equation}
Note that we use a relatively simple and straight-forward MLP model here to do opposite agent modeling. It has been shown effective in other studies like \cite{su2018discriminative}. We also carried out preliminary experiments on other more complicated designs such as \cite{weber2017imagination}. However, results have shown MLP's superior performance in our dialogue policy learning task.


\vspace{-1mm}
\subsection{Opposite Aware Q-Learning}
After obtaining the estimated opposite reaction $a_{t+1}^o{'}$, it serves as an extra input to the DQN-based policy component besides the original dialogue state representation $s_t$. Therefore, for the state and action composed deep Q-function $Q(s_t, a_t)$, we form a new state representation $\hat{s_t}$ as below:
\begin{equation}
\hat{s_t} = [s_t, E^oa_{t+1}^o{'}],
\end{equation}
in which $E^oa_{t+1}^o{'}$ introduces the knowledge of opposite agent into our policy learning. $E^o$ is the opposite action embedding matrix which maps the action into specific vector representation.
Given the output $a_t$ of $\argmax_{a'} Q(\hat{s_t}, a')$, the agent chooses an action to execute using an $\epsilon$-greedy policy that selects a random action with probability $\epsilon$ or otherwise follows the output $a_t$. We update the Q-function by minimizing the mean-squared loss function, which is defined as
\vspace{-1mm}
\begin{align}
\mathcal{L}_1(\theta) &= \mathbb{E}_{(s,a,r,s'){\sim}\mathcal{D}^L}[(y_i-Q(s,a))^2],\\
\label{eq:dqn}
y_i&=r+{\gamma}\max_{a'}Q'(s',a'),
\end{align}
\vspace{-1mm}
where $\gamma \in [0,1]$ is a discount factor, $\mathcal{D}^L$ is the replay buffer and $y_i$ represents the expected reward computed based on the transition.

\vspace{-1mm}
\subsection{Target Action Sampling}
\label{fake_sample}
In this subsection, we explain how the action $\hat{a_t}$ is sampled utilizing the above modules. For generating the true target action $a_t$, we predict it using a deep Q-network which takes as input an estimated opposite action $a_{t+1}^o{'}$ and the dialogue state $h_t^e$. However, we cannot get $a_{t+1}^o{'}$ without $\hat{a_t}$. Therefore, we further leverage this Q-network at hand. Specifcally, we feed an constant opposite action placeholder $a^o$ to the Q-function:
\vspace{-1mm}
\begin{equation}
\hat{a_t} = \argmax_{a{'}}Q([h_t^e, E^oa^o], a{'})
\end{equation}
\vspace{-1mm}
where $a^o$ serves as a constant opposite action. In our experiment, $a^o$ corresponds to the general actions which do not influence business logic, such as \textit{Hello} and \textit{Thanks}.

%

\vspace{-1mm}
\subsection{Action Regularization with Decay}
In our method, $\hat{a_t}$ is sampled from a distribution. At the very beginning of training, since the model is not well trained, the sampled $\hat{a_t}$ may perform badly, which would lead to slow convergence. Therefore, we apply action regularization to mitigate the difference between $\hat{a_t}$ and real $a_t$.
As the training progress goes on, such guidance becomes less effective, and we hope to encourage the model to explore more in the action space. Therefore, we adopt a decay mechanism inspired by \cite{zhang2019bridging}. The regularization term is defined as the cross entropy of $\hat{a_t}$ and real action:
\vspace{-1mm}
\begin{equation}
\label{eq:ar}
\mathcal{L}_2(\theta) = -\beta\sum_t a_tlog(\hat{a_t}),
\end{equation}
\vspace{-1mm}
where $\beta$ is the decay coefficient. The value of $\beta$ decreases along with time by applying a discount factor $\gamma$ in each epoch. As a consequence, a strict constraint on the sampled action is applied to avoid large action sampling performance drop at the beginning stage. After that, the constraint is continuously relaxed so that the model can explore more actions for better strategy.

To sum up, the final loss function for training our OPPA model is the weighted sum of the DQN loss and action regularization loss:
\vspace{-1mm}
\begin{equation}
\mathcal{L}(\theta) = w_1\mathcal{L}_1(\theta) + w_2\mathcal{L}_2(\theta).
\end{equation}
\vspace{-1mm}
%
%

\subsection{Reward Function}
When a dialogue session is completed, what we get are several dialogue acts or natural language utterances (when paired with NLU and NLG).
For most goal-oriented dialogues, the reward signal can be obtained from the user simulator or real user ratings.
However, when that reward is not available, an output prediction model is required which takes as input the whole dialogue session $X=\{x_1^s, x_2^s, ..., x_n^s\}$ where $X$ is a sequence of tokens, and outputs structured result to calculate the reward.

We use a bi-directional GRU model with an attention mechanism to learn a summarization $h^s$ of the whole session:
\begin{align}
h_j^o&={\rm BiGRU}(h_{j-1}^o, [Ex^s_j, h_j]),\\
h_j^a&=W^a[{\rm tanh}(W^hh_j^o)], \\
{\alpha}_j&=\frac{{\rm exp}(w{\cdot}h_j^a)}{{{\sum}_{t'}^{}}{\rm exp}(w{\cdot}h^a_{j'})}, \\
h^s&={\rm tanh}(W^s[h^g,{{\sum}_j}{\alpha}_jh_j]).
\end{align}
Note that in this process, we concatenated all the utterances by time order, and the subscript $j$ indicates the index of word in the concatenated sequence.
In addition, there may be multiple aspects of the output. For example, in a negotiation goal-oriented dialogue with multiple issues (we denote the book or hat items to negotiate on as issues), 
we need to get the output of each issue to calculate the total reward. Therefore, for each issue $o_i$, a specific softmax classifier is applied:
\begin{equation}
p_{\theta}(o_i|x_{0...T},g)={\rm softmax}(W^{o_i}h^s).
\end{equation}

After the structured output is predicted, we can obtain the  final reward by applying the task-specific reward function on the output.
\begin{equation}
r=f^R(o_1, o_2, ..., o_{N_o}),
\end{equation}
where $N_o$ is the number of output aspects and $f^R$ is the reward function which is often manually defined according to the task.

\section{Experiment}
Depending on the task, dialogues can be divided into cooperative and competitive ones. In a cooperative task, the aim can be reducing unnecessary interactions by inferring the opposite person's intention. While in competitive tasks, the aim is usually to maximize their own interests by considering the opposite agents' possible reactions. To test our method's wide suitability, we evaluated it on both cooperative and competitive tasks.

\vspace{-0.2cm}
\subsection{Dataset}
\label{sec:dataset}

For the cooperative task, we used MultiWOZ \cite{budzianowski2018multiwoz}, a large-scale linguistically rich multi-domain goal-oriented dialogue dataset, which contains 7 domains, 13 intents and 25 slot types.
There are 10,483 sessions and 71,544 turns, which is at least one order of magnitude larger than previous annotated task-oriented dialogue dataset.
Among all the dialogue sessions, we used 1,000 each for validation and test. Specifically, in the data collection stage, the user follows a specific goal to converse with the agent but is encouraged to change his/her goal dynamically during the session, which makes the dataset more challenging.

For the competitive task, we used a bilateral negotiation dataset \cite{lewis2017deal}, where there are 5,808 dialogues from 2,236 scenarios. In each session, there are two people negotiating to divide some items, such as books, hats and balls. Each kind of item is of different value to each person, thus they can give priority to valuable items in the negotiation. For example, a hat may worth $5$ for person $A$ and $3$ for person $B$, so $B$ can give up some hat in order to get other valuable items.
To conduct our experiment, we further labeled the dataset with system dialogue actions.

\subsection{Experimental Settings}
\label{sec:train_detail}
We implemented the model using PyTorch \cite{paszke2017automatic}.
The hyper-parameters were decided using validation set.
The dimension of ${\rm GRU}_o$ hidden state is 256, and the hidden state size of ${\rm GRU}_g$ and ${\rm GRU}_w$ are 64 and 128 respectively. The size of $h_s$ is 256. As for the Q-function, the size of $s_t$ is 256. $\epsilon$-greedy is applied for exploration. The buffer size of $D$ is set to 500 and the update step $C$ is 1.

Note that due to the complexity of MultiWOZ, the error propagation problem caused by NLU and NLG is serious. Therefore, the cooperative experiment is conducted on the dialogue act level. In the experiment, our proposed model interacts with a robust rule-based user simulator, which appends an agenda-based model \cite{schatzmann2007agenda} with extensive manual rules. The simulator gives user response, termination signal and goal-completion feedback during training.
For the competitive task, the experiment is on natural language level. Following \cite{lewis2017deal}, we built a seq2seq language model for the NLU and NLG module, which is pre-trained on the negotiation corpus. 

Our proposed model was first pre-trained with supervised learning (SL). Specifically, we pre-trained the opposite estimator $\pi_o$ and the Q-function $Q(s, a)$ via supervised learning and imitation learning. 
We then fine-tuned the model using reinforcement learning (RL). 
The reward of the MultiWOZ experiment consists of two parts:  a) a small negative value in each turn to encourage shorter sessions and b) a large positive reward when the session ends successfully. Note that the task completion signal is obtained from the user.
For the negotiation experiment, the reward is the total value of item items that the agent finally got. In the negotiation dataset, the reward is given by the proposed output model described in the Reward Function section.


\subsection{Baselines}
\label{sec:basline}
To demonstrate the effectiveness of our proposed model, we compared it with the following baselines. 
For the MultiWOZ task, we compared with:
\vspace{-0.2cm}
\begin{itemize}[leftmargin=*]
    \setlength{\itemsep}{1pt}
    \item \textbf{DQN}: The conventional DQN \cite{mnih2015human} algorithm with a 2-layer fully-connected network for Q-function.\vspace{-0.2cm}
    \item \textbf{REINFORCE}: The REINFORCE algorithm \cite{williams1992simple} with a 2-layer fully-connected policy network.\vspace{-0.2cm}
    \item \textbf{PPO}: Proximal Policy Optimization \cite{schulman2017proximal}, a policy-based RL algorithm using a constant clipping mechanism.\vspace{-0.2cm}
    \item \textbf{DDQ}: The Deep Dyna-Q \cite{peng2018deep} algorithm which introduced a world-model for RL planning.
\end{itemize}
\vspace{-0.2cm}
Note that the DQN can be seen as our proposed model without opposite estimator  (\textbf{OPPA w/o OBE}).
For the negotiation task, we compared with:
\vspace{-0.2cm}
\begin{itemize}[leftmargin=*]
    \setlength{\itemsep}{1pt}
    \item \textbf{SL RNN}: A supervised learning method that is based on an RNN language generation model.\vspace{-0.2cm}
    \item \textbf{RL RNN}: The reinforcement learning extension of SL RNN by refining the model parameters after SL pretraining.\vspace{-0.2cm}
    \item \textbf{ROL}: SL RNN with goal-based decoding in which the model first generates several candidate utterances and chooses the one with the highest expected overall reward after rolling out several sessions.\vspace{-0.2cm}
    \item \textbf{RL ROL}: The extension of RL RNN with rollout decoding.\vspace{-0.2cm}
    \item \textbf{HTG}: A hierarchical text generation model with planning \cite{yarats2017hierarchical}, which learns explicit turn-level representation before generating a natural language response.
\end{itemize}
\vspace{-0.2cm}
Note that the rollout mechanism used in ROL and RL ROL also endows them with the ability of ``seeing ahead'' in which the candidate actions' rewards are predicted using a random search algorithm, while our OPPA explicitly models the opposite's behavior.
RL RNN, RL ROL and HTG used the REINFORCE \cite{williams1992simple} algorithm for reinforcement learning on both strategy and language level, while in OPPA we used the DQN \cite{mnih2015human} algorithm only on strategy level.
To further examine the effectiveness of our proposed action regularization with decay, we did an ablation study by removing the regularization with decay part in Equation \ref{eq:ar} (\textbf{OPPA w/o A}).


\subsection{Evaluation Metric}
For the evaluation of experiments on MultiWOZ, we used the number of turns, inform F1 score, match rate and success rate.
The \textbf{Number of turns} is the averaged number on all sessions, and less turns in cooperative goal-oriented task can promote user satisfaction. 
\textbf{Inform F1} evaluates whether all the slots of an entity requested by the user has been successfully informed.
We use F1 score because it considers both the precision and recall so that a policy which greedily informs all slot information of an entity won't get a high score.
\textbf{Match rate} evaluates whether the booked entities match the goals in all domains. The score of a domain is 1 only when its entity is successfully booked.
Finally, a session is considered \textbf{successful} only if all the requested slots are informed (recall = 1) and all entities are correctly booked.

For the negotiation task, we used the averaged scores (total values of items) of all the sessions and those with an agreement as the primary evaluation metrics following \cite{lewis2017deal}. The percentage of agreed and Pareto optimal\footnote{A dialogue is Pareto optimal if neither agent’s score can be improved without lowering the other's score.} sessions are also reported.

 \vspace{-0.2cm}
\begin{table}[!htp]
    \center
    \small
    \begin{tabular}{c  c  c  c  c}
        \thickhline
        Method & \#Turn & Inform F1 & Match & Success \\
        \hline
        DQN & 10.50 & 78.23 &  60.31 & 51.7\\
        REINFORCE & 9.49 & 81.73 & 67.41 & 58.1\\
        PPO  & 9.83 & 83.34 & 69.09 & 59.0\\
        DDQ & 9.31 & 81.49 & 63.10 & 62.7\\
        \hline
        OPPA w/o A & 8.19 & 88.45 & 77.18 & 75.2\\
        OPPA & 8.47 & 91.68 & 79.62 & 81.6\\
        \hline
        \textit{Human} & 7.37 & 66.89 & 95.29 & 75.0 \\
        \thickhline
    \end{tabular}
    \vspace{-0.2cm}
    \caption{\label{tab:task_oriented} The results on MultiWoZ dataset, a large scale multi-domain task-oriented dialog dataset. We used a rule-based method for DST and Agenda-based user simulator. The DQN method can be regard as OPPA w/o OBE. Human-human performance from the test set serves as the upper bound.}
    \vspace{-0.2cm}
\end{table}

\subsection{Cooperative Dialogue Analysis}

The results on MultiWOZ dataset are shown in Table \ref{tab:task_oriented}.
OPPA shows superior performance on task success rate than other baseline methods due to the considerable improvement in Inform F1 and Match rate.
By first infer the next action of the opposite agent, the target agent policy can make better choices to match the reward signal during training.
When compared with human performance, OPPA even achieves a higher success rate, although the number of turns is still higher. This might be due to the fact that the user is sensitive to the dialogue length. When a dialogue becomes intolerably long, many user will leave without completing the dialogue.
By taking actions in account of the inferred opposite action, the target agent can also make the dialogue more efficiently by avoiding some lengthy interactions, which is extremely important in applications where the user is sensitive to dialogue length. 

Meanwhile, DDQ achieves higher task success rate than other baseline models since it also models the behavior of opposite agent through world model. However, it makes use of the learned world model by providing more simulated experiences, which does not give a direct hint on how to act in the middle of a session.
Therefore, in its experiments, it still gets longer dialogue sessions and a lower success rate than OPPA.

If we remove the action regularization mechanism, we can see an obvious decline on performance, which is as expected. The action regularization is introduced to mitigate the difference between sampled $\hat{a_t}$ and real $a_t$, so there can be a large discrepancy between the sampled and real actions if we remove it at the early training stage.
When the $\hat{a}t$ is not reliable, the consequent estimated opposite action $a_{t+1}'$ also becomes noisy, which leads to performance drop.


\begin{table*}[!htp]
    \small
    \setlength{\leftskip}{-8pt}
    \begin{tabular}{c  c c  c c c c  c c}
        \thickhline
        \multirow{2}{*}{Method} & \multicolumn{2}{c}{vs. SL RNN} & \multicolumn{2}{c}{vs. RL RNN} & \multicolumn{2}{c}{vs. ROL} & \multicolumn{2}{c}{vs. RL ROL} \\
        \cline{2-9}
        & All & Agreed & All & Agreed & All & Agreed & All & Agreed \\
        \hline
        SL RNN & 5.4 vs. 5.5 & 6.2 vs. 6.2 & -& -& -& -& -&- \\
        RL RNN & 7.1 vs. 4.2 & 7.9 vs. 4.7 & 5.5 vs. 5.6 & 5.9 vs. 5.8 & -& -&- &- \\
        ROL & 7.3 vs. 5.1 & 7.9 vs. 5.5 & 5.7 vs. 5.2 & 6.2 vs. 5.6 & 5.5 vs. 5.4 &5.8 vs. 5.9 &- &- \\
        RL ROL & 8.3 vs. 4.2 & 8.8 vs. 4.5 & 5.8 vs. 5.0 & 6.5 vs. 5.5 & 6.2 vs. 4.9 & 7.0 vs. 5.4 & 5.9 vs. 5.8 & 6.4 vs. 6.3 \\
        HTG & 8.7 vs. 4.4 & 8.8 vs 4.5 & 6.0 vs. 5.1 & 6.9 vs. 5.5 & 6.5 vs. 5.0 & 6.9 vs. 5.3 & 6.5 vs. 5.6 & 7.0 vs. 6.3\\
        \hline
        OPPA w/o OE & 8.2 vs. 4.2 & 8.8 vs. 4.7 & 6.1 vs. 5.2 & 6.8 vs. 5.6 & 6.5 v.s. 4.8 & 7.0 v.s. 5.3 & 5.7 v.s. 5.8 & 6.5 v.s. 6.4\\
        OPPA w/o A & 8.7 vs. 4.1 & 8.9 vs. 4.3 & 6.3 vs. 5.0 & 7.2 vs. 5.4 & 6.5 vs. 4.8 & 7.2 vs. 5.4 & 6.5 vs. 6.1 & 7.1 vs. 6.8\\
        OPPA & 8.8 vs. 3.9 & 9.0 vs. 4.1 & 6.7 vs. 4.6 & 7.3 vs. 5.2 & 6.8 vs. 4.2 & 7.4 vs. 5.1 & 6.7 vs. 6.0 & 7.2 vs. 6.6  \\
        \thickhline
    \end{tabular}
    \vspace{-0.2cm}
    \caption{\label{tab:nego_score} The results of our proposed OPPA and the baselines on the negotiation dataset. \textit{All} and \textit{Agreed} indicates averaged scores for all sessions and only the agreed sessions respectively.}
\end{table*}

\begin{table*}[!htp]
    \centering
    \small
    \begin{tabular}{c  c c  c c c c  c c}
        \thickhline
        \multirow{2}{*}{Method} & \multicolumn{2}{c}{vs. SL RNN} & \multicolumn{2}{c}{vs. RL RNN} & \multicolumn{2}{c}{vs. ROL} & \multicolumn{2}{c}{vs. RL ROL} \\
         \cline{2-9}
        & Agreed(\%) & PO(\%) & Agreed(\%) & PO(\%) & Agreed(\%) & PO(\%) & Agreed(\%) & PO(\%) \\
        \hline
        SL RNN & 87.9 & 49.6 &- & -&- & -&- &- \\
        RL RNN & 89.9 & 58.6 & 81.5 & 60.3 &- &- &- & -\\
        ROL & 92.9 & 63.7 & 87.4 & 65.0 & 85.1 & 67.3 &- & -\\
        RL ROL & 94.4 & 74.8 & 85.7 & 74.6 & 71.2 & 76.4 & 67.5 & 77.2 \\
        HTG & 94.8 & 75.1 & 88.3 & 75.4 & 83.2 & 77.8 & 66.1 & 73.2 \\
        \hline
        OPPA w/o OBE & 94.6 & 74.6 & 87.9 & 75.2 & 79.3 & 78.2 & 73.7 & 77.9 \\
        OPPA w/o A & 95.6 & 77.9 & 91.9 & 77.4 & 82.4 & 78.8 & 78.0 & 79.5 \\
        OPPA & 95.7 & 77.7 & 91.4 & 77.2 & 82.3 & 79.1 & 78.2 & 79.7\\
        \thickhline
    \end{tabular}
    \vspace{-0.2cm}
    \caption{\label{tab:nego_per} The proportion of agreed and Pareto optimal (PO) sessions for our proposed OPPA and the baselines on the negotiation dataset.}
    \vspace{-0.3cm}
\end{table*}

\vspace{-0.2cm}
\subsection{Competitive Dialogue Analysis}

Table \ref{tab:nego_score} shows the scores for all sessions and for only agreed ones.
When comparing with the seq2seq models, OPPA achieves significantly better results. This can be attributed to the hierarchical structure of OPPA. The sequence models only take as input (and outputs) the word-level natural language utterances, without explicitly modeling turn-level dialogue actions. In this way, the parameters for linguistic and strategy functions are tangled together, and the back-propagation errors can influence both sides.
As for the two ROL models, although they can predict the value of a candidate action in advance, they still cannot beat OPPA. The reason is that the rollout method did not explicitly maintain an estimation of the opposite agent as our OPPA did. Instead, it just estimates the candidate acitons' rewards based on Monte Carlo search by using its own model for predicting future movements. Therefore, when the opposite model's behavior is not very familiar to the target agent, 
the estimated reward becomes unreliable.

The HTG model also used a hierarchical framework by learning an explicit turn-level latent representation. By doing this, it obtains higher scores than the seq2seq models. However, it does not make any assumptions about the opposite agent. Therefore, its scores are still lower than OPPA, although the discrepancy narrows down.

By removing the opposite estimator, we find that the performance of OPPA w/o OBE drops significantly compared to that of OPPA. This ablation study directly verifies the effectiveness of our proposed opposite behavior estimator. There fore, modeling the opposite policy in one's mind is a crucial source to achieve better results in competitive dialogue policy learning.

When comparing with OPPA w/o A which removed action regularization, we can see that the OPPA model gets better results. This verifies the importance of regularizing the action sampling. By controlling the difference between real and model generated actions, we can keep the opposite model consistent with the real opposite agent at the early training stage.

The percentage of agreed and Pareto optimal session are shown in Table \ref{tab:nego_per}. As we can see, the percentage of Pareto optimal increases in our method, showing that the OPPA model can explore the solution space more effectively. However, the agreement rate decreases when the opposite model gets stronger. This phenomenon is also found in \cite{lewis2017deal} when they change the opposite agent from SL RNN to real human. This can be attributed to the aggressiveness of the agent: when both agents act aggressively, they are less likely to reach an agreement. The SL RNN model simply imitates the behavior in the dialogue corpus, while the ROL and RL mechanisms both help the agent to explore more spaces, which makes them more aggressive on action selection.

\vspace{-0.2cm}
\subsection{Human Evaluation}
To better validate our propositions, we further conducted human evaluation by making our model conversing with real user. 
We only conducted human evaluation on the negotiation task since the MultiWOZ model is implemented on the dialogue act level.
We tested the models on a total of 1,000 dialogue sessions. In the evaluation, the users conversed with the agent, and the total item values are used as the evaluation metric. 
The results are shown in Table \ref{tab:human_nego}. We can see that our proposed OPPA outperforms the baseline models. The system score are lower than that in Table \ref{tab:nego_score}, and the discrepancy between \textit{All} and \textit{Agreed} results is large.
This can be due to the high intelligence and aggressiveness of real humans who want to get as more values as possible and do not make compromises easily. Due to this reason, the sessions become considerably lengthy, and the target agent exceeds our length limit before reaching an agreement.

\begin{table}[!htp]
    \center
    \small
    \begin{tabular}{c  c  c }
        \thickhline
        Method & All & Agreed \\
        \hline
        RL ROL & 4.5 vs. 5.2 & 7.8 vs. 7.1 \\
        HTG & 4.8 vs. 4.7 & 8.0 vs. 7.2\\
        OPPA w/o A & 4.7 vs. 4.9 & 8.4 vs. 6.7\\
        OPPA & 5.2 vs. 5.1 & 8.2 vs. 6.5 \\
        \thickhline
    \end{tabular}
    \vspace{-0.2cm}
    \caption{\label{tab:human_nego} The rewards of each model vs. human user.}
    \vspace{-0.3cm}
\end{table}


	\vspace{-0.2cm}
\section{Conclusion}
In this work, we present an opposite agent-aware dialogue policy model which actively estimates the opposite agent instead of doing passive learning from experiences. 
We have shown that it is possible to harvest a reliable model of the opposite agent through more efficient dialogue interactions. By incorporating the estimated model output as part of dialogue state, the target agent shows significant improvement on both cooperative and competitive goal-oriented tasks.
As future work, we will explore multi-party dialogue modeling in which multi-agent learning techniques can be applied.
	

	\bibliography{acl2020}
	\bibliographystyle{acl_natbib}
	
\end{document}


\section*{Appendix A: Dialog Example}
\label{sec:supplemental}

\begin{table}[ht]
	\footnotesize
	\small
	\setlength{\leftskip}{-100pt}
	\begin{tabular}{l l| l l}
		\hline
		\textbf{Input} &  & \textbf{Input} &  \\
		PLOBE  & 3xbook value=0 1xhat value=6 1xball value=4 & PLOBE  & 1xbook value=7 3xhat value=1 3xball value=0\\
		RL+ROL  & 3xbook value=1 1xhat value=4 1xball value=3 & RL+ROL  & 1xbook value=7 3xhat value=0 3xball value=1\\
		\hline
		RL+ROL  & i would like the hat and the ball . & PLOBE  & i would like the book and 2 hats .\\
		PLOBE & i need the ball and the hat & RL+ROL & i need the book and the balls . \\
		RL+ROL & i really need the hat , but you can have the rest & PLOBE &  i need the book and the 1 hat \\
		PLOBE & i need the hat and the ball & RL+ROL & i need the book and 1 ball \\
		RL+ROL & okay & PLOBE & okay , i'll take the book and 1 hat . \\
		\hline
		\textbf{Output} & ~~~~~~~~~~~~~~~~~~~~~~~~~~~~~~~~~~~~~~~~~~~~ \textbf{Reward} & \textbf{Output} & ~~~~~~~~~~~~~~~~~~~~~~~~~~~~~~~~~~~~~~~~~~~~ \textbf{Reward} \\
		PLOBE & 1xhat 1xball ~~~~~~~~~~~~~~~~~~~~~~~~~~~~10/10 & PLOBE & 1xbook 2xhat ~~~~~~~~~~~~~~~~~~~~~~~~~~~9/10 \\
		RL+ROL & 3xbook ~~~~~~~~~~~~~~~~~~~~~~~~~~~~~~~~~~~ 3/10 & RL+ROL & 2xhat 3xball ~~~~~~~~~~~~~~~~~~~~~~~~~~~~ 3/10 \\
		\hline
	\end{tabular}
	\caption{Example dialogs between PLOBE agent and RNN+ROL agent. One example starts with the RL+ROL agent and the other example starts with the other agent.}
\end{table}

\section*{Appendix B: Human Evaluation}
The human evaluation experiment is conducted on the Amazon Mechanism Turk (AMT) platform, in which the human users were randomly paired with the agents. We ran 1,800 dialog sessions for the three datasets (Negotiation, DSTC2 and Fashion) and four models (RL ROL, HTG, PLOBE- and PLOBE).

The users were shown an example dialog and some instructions to make them act properly. In each conversation, the users were firstly shown some context information, and then the conversation started.
In the negotiation dialog, the context information describes the item values and amount for the human user.
For the task-oriented dialog, the context information is the user's goal. More specifically, for the DSTC2 dataset, it describes the values for a specific restaurant, and the slots to be requested; for the Fashion dataset, it describes the values of a specific item.